\documentclass[twoside,10pt]{article}

\usepackage{graphicx}
 \usepackage[utf8]{inputenc} 
 \usepackage[T1]{fontenc}    

\usepackage{url}            
\usepackage{booktabs}       
\usepackage{amsfonts}       
\usepackage{nicefrac}       

\usepackage{geometry}
\usepackage{wrapfig}
\usepackage{nicefrac}

\usepackage{todonotes}

\usepackage{jmlr2e}
\usepackage{amsmath}
\usepackage{bbding}
\usepackage{enumitem}
\usepackage{mathtools}

\newcommand{\R}{\ensuremath{\mathbb{R}}}

\newcommand{\Model}{\ensuremath{M_\theta}}

\newcommand{\Cont}{\ensuremath{{C(\R)}}}
\newcommand{\Alg}{\ensuremath{\mathfrak{L}}}

\ShortHeadings{A Foliated view of Transfer Learning}{Petangoda, Monk and Deisenroth}
\firstpageno{1}

\begin{document}
      
\title{A Foliated View of Transfer Learning}

\author{\name Janith Petangoda \email jcp17@ic.ac.uk \\
       \addr Department of Computing\\
	   Imperial College London\\
	   London, United Kingdom
       \AND
       \name Nicholas A. M. Monk \email n.monk@sheffield.ac.uk \\
       \addr School of Mathematics and Statisitics\\
	   University of Sheffield\\
	   Sheffield, United Kingdom
	   \AND
	   \name Marc Peter Deisenroth \email m.deisenroth@ucl.ac.uk \\
       \addr Department of Computer Science\\
	   University College London\\
	   London, United Kingdom}

\editor{}

\maketitle

\begin{abstract}
Transfer learning considers a learning process where a new task is solved by transferring relevant knowledge from known solutions to \emph{related tasks}. While this has been studied experimentally, there lacks a foundational description of the transfer learning problem that exposes what related tasks are, and how they can be exploited. In this work, we present a definition for relatedness between tasks and identify foliations as a mathematical framework to represent such relationships. 
\end{abstract}

\section{Introduction} \label{sec:introduction}
As the complexities of the learning problems we want to solve using machine learning (ML) increase, so too must our solutions; efficient algorithms to find such solutions are imminently necessary. Transfer learning (TL), where knowledge is transferred from existing solutions to new problems provides an answer to this. TL asks if learned models can aid in the learning of new models; can past experience make learning easier and better? How can we exploit \emph{structural similarities between problems} to learn more generally and holistically? This is further motivated by our own experiences as life-long learning agents. To this end, we must be able to deliberately design models and algorithms that can do this; a foundational understanding of TL is paramount. Presently, we will attempt to make fundamental strides in that direction.

\begin{figure}[ht] 
    \centering
    \includegraphics[width=0.37\textwidth]{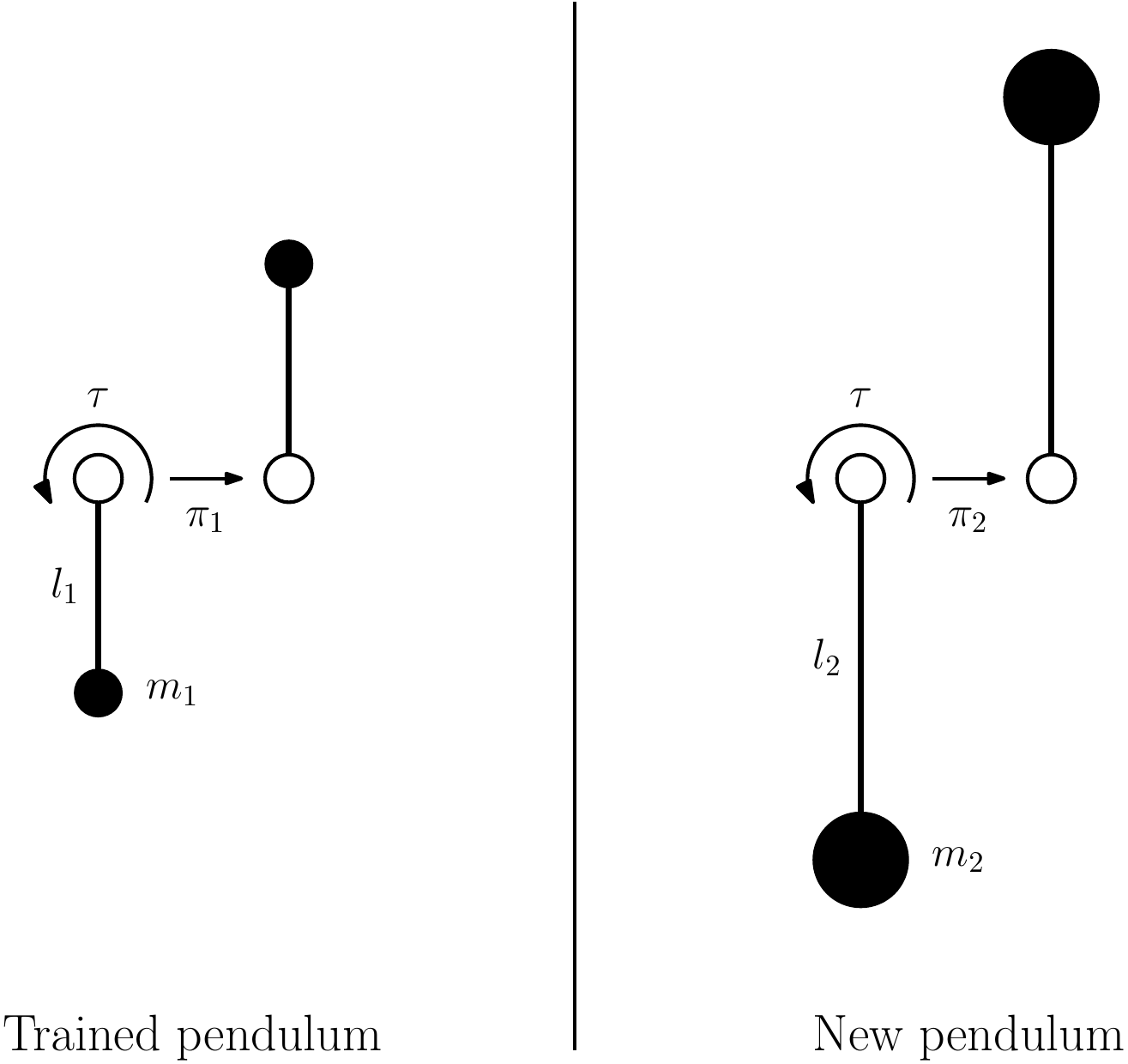}
    \caption{Transfer-learning problem; this example particularly applies to reinforcement learning. Having trained a policy $\pi_1$ that swings up and balances a pendulum with mass $m_1$ and length $l_1$ (left pane), we want to learn a policy $\pi_2$ for a new pendulum with properties $m_2$ and $l_2$ (right pane).}
    \label{fig:pendulum_example}
\end{figure}

Let us introduce the TL problem by considering two pendulums (see Figure \ref{fig:pendulum_example}) as a guiding example. These pendulums are similar in all aspects except for their masses and lengths. Suppose that we had solved some ML task on one of the pendulums using a learning algorithm; this task could, for example, be to learn a control policy using reinforcement learning (RL). Given this solution, we ask how we should complete the corresponding task on the second pendulum.

One option is to naively run the same learning algorithm on the second problem. Under mild assumptions, we will obtain a solution to the new problem consuming similar computational resources. However, we know that these two pendulums are \emph{related}; they share the dynamical properties of pendulums. Therefore, we would expect that the solutions to the learning tasks are \emph{similarly related}. We should not have to solve the second problem from scratch; instead, we should exploit the shared structure between the problems to inform the solution to the new task. 

Similar scenarios can be found in other applications of ML. For example, learning to walk and then being asked to run, learning to classify images from a particular point of view and then being asked to classify images from a different angle, and learning to transcribe an accent of a language and then being asked to transcribe a different accent of the same language; TL is useful in many applications \citep{Pan2009}. All these problems characteristically contain some relationship between the tasks that are to be solved. Answering the question of \emph{what the common information} between related tasks is, and \emph{how} to exploit it is then fundamental to transfer learning. This common information can be regarded as shared \emph{structure} between the tasks. For example, in the case of the two pendulums, where the dynamics are similar, this shared structure is the \emph{form} of the ODEs that can model their behaviours. 

TL was initially considered in the human-inspired lifelong learning problem \citep{Thrun1995, Thrun1996}, which has continued on in Continual Learning \citep{Parisi2017, Parisi2019}. A modern take on transfer learning can be found in the study of meta-learning \citep{Hospedales2020}, and few-shot learning \citep{Chen2019}. Meta-learning includes work such as \citep{Finn2017,Steindor2018, Grant2018, Hausman2018, Janith2019}, which consider learning higher-level `meta' properties that are common amongst a group of related tasks. This is also carried out in few-shot learning methods, such as \citep{Snell2017, Vinayls2016}, which aim to explicitly learn new tasks in a \emph{few} learning iterations. Of particular note is the  recent work by \citep{Zhou2012}, which describes meta-learning symmetries shown by the data. 

These modern works are practical in nature, and explore experimentally useful techniques; they however lack a formal foundation. They cannot meaningfully answer the question of \emph{what} the common structure between related tasks is, and how it is exploited. Knowledge of such relationships will allow for a deeper understanding of \emph{what} happens during TL, \emph{where} it is useful, and \emph{why} it works as it does. For example, \citep{Tian2020} suggests that meta-learning algorithms might not be working as well as we think, but we cannot yet prove why. In this paper, we provide a formalism for Transfer Learning that allows one to study such structure mathematically using the theory of foliations. 

Sections \ref{sec:representations} and \ref{sec:relatedness} will intuitively introduce our formalism by describing its key component ideas. They provide a formal way of describing ML problems, and how to consequently think about relatedness between tasks, respectively. In Section \ref{sec:formalism}, we present our formalism more rigorously. Section \ref{sec:future} will end with some thoughts on how this formalism is used implicitly in existing work, and possible future applications. 

\section{Representations} \label{sec:representations}
Representations\footnote{This should be distinguished from representation learning as described in \citep{Bengio2013} and similar work, where a representation of data that extracts useful information is learned. A representation in our sense can provide this, but is a more general notion, which can apply to an arbitrary set of things.} are a way of describing machine learning problems and models. Intuitively, a representation is a mechanism by which we can \emph{describe} and \emph{realise} abstract objects \citep{Marr1982} in terms of a scheme that uses fundamental building blocks and rules. They allow us to communicate, manipulate and reason about one or more such abstract objects. For example, the abstract object we usually denote by $42$ is an abstract object with certain  properties, such as being proceeded by $43$ and proceeding $41$. We can also equivalently, and without loss of information, represent it as forty-two in English,  
 and $101010$ in binary. A representation is not unique; we can describe the same object in different ways. It is not necessarily exact; natural numbers can approximate real numbers for example. A representation can approximate abstract objects up to some desired properties. A representation must also reflect axiomatic assumptions and properties regarding our abstract objects.
 
A ML problem can be cast as finding an approximate description of a target function such that generalisation is possible. This is clearest in supervised learning, where using pairs of inputs and outputs from a target function, we are tasked with finding a suitable proxy for it. As an example consider the set of continuous scalar valued functions on $\R$, denoted by $\Cont$. Suppose from such an element $f$, we are given $n$ input/output pairs, giving us a data set $D_f = \{(x_i, y_i)\}_{i=1}^n$. Here, $y_i$ are the observed values of $f(x_i)$ under some noise model. Then, we would assume a class of models $\Model$, which could for example, be a class of neural networks (NNs) of a chosen architecture. A particular NN is identified by the values of its weights and biases, denoted here by $\theta$. Given these, we would find the model $m_f \in \Model$ that best approximates the function $f$ using a learning algorithm, which includes an appropriate loss function to assess suitability.

In this way, the learning algorithm is a process of finding a good and useful description of an abstract object under a representation scheme from the data. That is, if the learning algorithm is denoted by $\Alg_L$, then for the example above, it can be described as a map 
\begin{equation}
\Alg_L:\Cont \rightarrow M_\theta.	
\end{equation}
In practice, the input space is the data set; the map $\Alg_L$ above is the limit as $n \rightarrow \infty$. 

The resulting approximating model is then a description of the target function with respect to the representation scheme given by the NNs. Another approximate representation of the target function is the data itself. However, there is no notion of generalisation here; the data alone does not allow us to query the output values at new input points. We see here that we have placed structure on our chosen representation scheme such that our desired goal of generalisability is satisfied.  

What then, are the necessary structural requirements for a suitable representation for TL? Firstly, we can note that TL, by its nature, considers at least two tasks; the two pendulum tasks for example. We could choose two distinct representations for each task, but we are interested in exploiting some relationship between the tasks. It is then reasonable to assume that the representations themselves must be related; they must communicate the relationship between tasks. In the present work, we borrow from the literature of multi-task learning (MTL) \citep{Caruana1997, Ruder2017} and assume that the \emph{same} representation is used for all tasks we would like to consider transferring between.

The question we would now like to answer is what a particular solution to a task, under this chosen representation, tells us about the solution to a related task? To answer this, we must be precise about what we mean by relatedness. In the next section, we will provide a way to describe such relationships, and see that this gives rise to a geometric structure on the class of models that exactly represents them.

\section{Relatedness} \label{sec:relatedness}
The definition of relatedness that we will use borrows from \citep{Ben2003, Ben2008}. While these works declare relatedness probabilistically, we will be interpreting a similar definition geometrically. We will introduce structures that allow us to argue greater generality than our contemporaries. Additionally, we will draw a clear distinction between \emph{relatedness} and \emph{similarity}. In this section, we give an intuitive argument for how we consider relatedness; we will discuss a formal description in Section \ref{sec:formalism}.

Our notion of relatedness tries to tell us \emph{what} changes between tasks. The key to this is a set of transformations that can act on a set of tasks; each element of this set transforms one task to another. We can then define a \emph{relationship} between two tasks as an element of this transformation set. That is, two tasks are related if there exists a particular element of this set that can transform one task to another. A \emph{set of related tasks} then contains all tasks that can be transformed from one to another using an element of the considered transformation set. Such a set of related tasks is an equivalence class, where the equivalence relationship is defined by the transformation set. We will denote $[f]_\sim$ as the equivalence class (the set of related tasks that $f$ belongs to), where $\sim$ is the equivalence relation that characterises this equivalence class.

 Relatedness therefore depends on the chosen transformation set. We can imagine that there are various transformation sets that can be used on a set of tasks. As an example, consider the set of continuous functions $\Cont$ as we defined above. One set of transformations we can think of comprises vertical translations of functions by a constant value along the entire domain; let us denote this $\Pi_1$. That is, for $f \in \Cont$ and $\pi_a \in \Pi_1$, 
 \begin{equation}
 	\pi_a(f)(x) = f(x) + a,
 \end{equation}
  for $x \in \R$ and $a \in \R$. The set of transformations is parameterised by a single real number. A set of related functions will contain all functions that can be expressed as a vertical translation of another; for example, the set of all sinusoids of constant period and amplitude that are vertical translations of each other are in a set of related tasks. 

 We could also consider a set of transformations that include both vertical translations as well as vertical scalings; this will be $\Pi_2$. That is, for $f \in \Cont$ and $\pi_{ab} \in \Pi_1$, 
 \begin{equation}
 	\pi_{ab}(f)(x) = bf(x) + a,	
 \end{equation}
 for $x \in \R$ and $a, b \in \R$. The set of transformations is now parameterised by two real numbers. As before, sets of related functions can be created by collecting all functions that can be translated or stretched into one another. Compared to the former case, these sets of related functions are larger, since the set of transformations is larger. Thus, we can find a relationship between functions that were not related with respect to $\Pi_1$. Relatedness only makes sense in the context of the chosen transformation set; this set informs us of the expressiveness of a notion of relatedness. 
 
How is this notion of relatedness useful to transfer learning? We have described the single task supervised learning problem previously. Suppose we are then given data from another element $(g \in \Cont) \neq f$. We want to find the best approximating model for it from $\Model$; that is, an element $m_f \in \Model$ that best approximates $f \in \Cont$. Crucially, we are told that this new function is \emph{related} to $f$ in the sense of a set of transformations. Suppose that this set of transformations is $\Pi_1$. 

\begin{figure}
    \centering
    \includegraphics[width=0.6\textwidth]{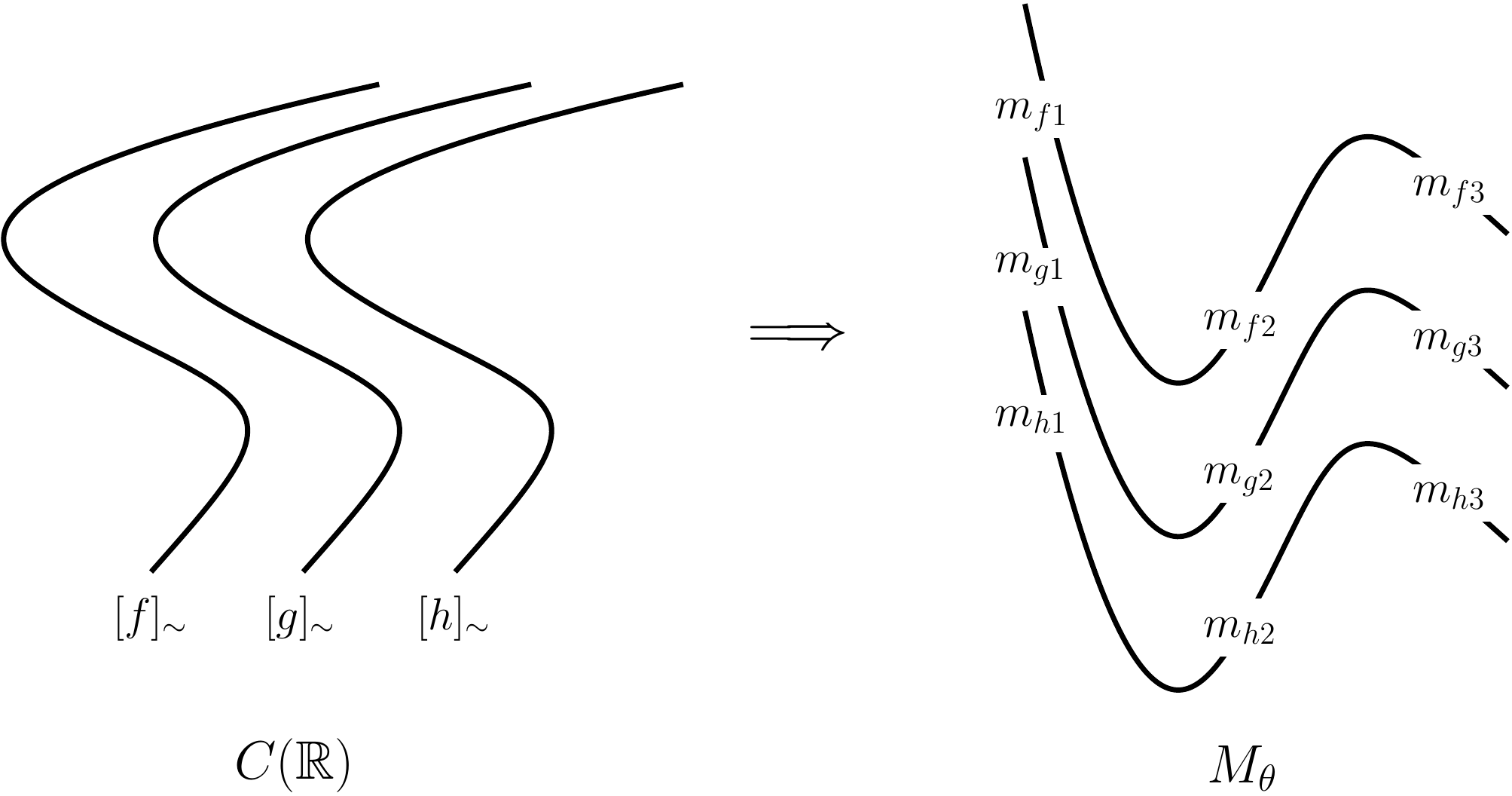}
    \caption{Equivariance between the set of continuous functions $\Cont$, and the set of approximators $\Model$ under $\Pi_1$. Along each line in $\Cont$, we have all elements that are related to each other by an  associated set of transformations; we would like that the models approximating target functions on a line to also be similarly related.}
    \label{fig:equivariance_c}
\end{figure}

As with the case of the pendulum problem from Section \ref{sec:introduction}, we could solve each of these tasks independently using the same learning algorithm we used to find $m_f$. In TL, we want to exploit the known relationship between tasks. We can assume that the best approximators $m_f$ and $m_g$ are also related \emph{in the same way} as $f$ and $g$. This is a statement of \emph{equivariance} of the learning algorithm $\Alg_L$. Equivariance tells us that a transformation in the domain of a map corresponds to a similar transformation of the output of that map. That is, if the original tasks are related in a particular way, then their approximators are related in a similar manner. We visualise this in Figure \ref{fig:equivariance_c} for $\Cont$ considered with $\Pi_1$ as the transformation set.

In Figure \ref{fig:equivariance_c}, each set of related tasks in $\Cont$ is drawn as a one-dimensional line. This is because, as said previously, each transformation in $\Pi_1$ can be  identified by a particular value of $\R$. In Section \ref{sec:invariants}, we will explain formally why this is appropriate. The desirable equivariance implies that in the space of approximating models $\Model$, the models that describe target functions that are from the same `line' in $\Cont$ should also lie on a line in the space of models. That is, a set of related tasks must surely induce a set of related approximating models.

We can also see that by considering the associated set of transformations, we partition the space of continuous functions into non-overlapping spaces; we call them the \emph{parallel spaces} or \emph{transfer spaces}. Each parallel space describes a set of related tasks or an equivalence class of tasks. Thus, a particular continuous function can be identified by the parallel space it lies on, and where on the parallel space it is (in terms of a coordinate system defined on the parallel space). Since the parallel space is of a lower dimension than the original space, knowing which parallel space a particular solution is on simplifies the problem of finding it. The dimensionality of the transformation set gives us the dimension of the parallel space \citep{Olver1995}. This dimensionality directly relates to how much easier learning of the new problem will be, as once the appropriate parallel space is known, we only have to search this lower-dimensional space for our desired solution.

The set of transformations and the statement of equivariance give us a good description of TL, and shows us how the key benefit of TL is achieved. That is, in transfer, we exploit some relational structure in the space of \emph{learning tasks}. This structure is described with respect to a set of transformations that gives us a precise notion of related tasks. The statement of equivariance then tells us that such relationships should be mimicked in the space of solutions to the learning tasks. These together describe a partitioning of both the task and solution spaces such that knowing which parallel space our task and solution exist on makes the search for the solution easier.

\begin{figure} 
    \centering
    \includegraphics[width=0.7\textwidth]{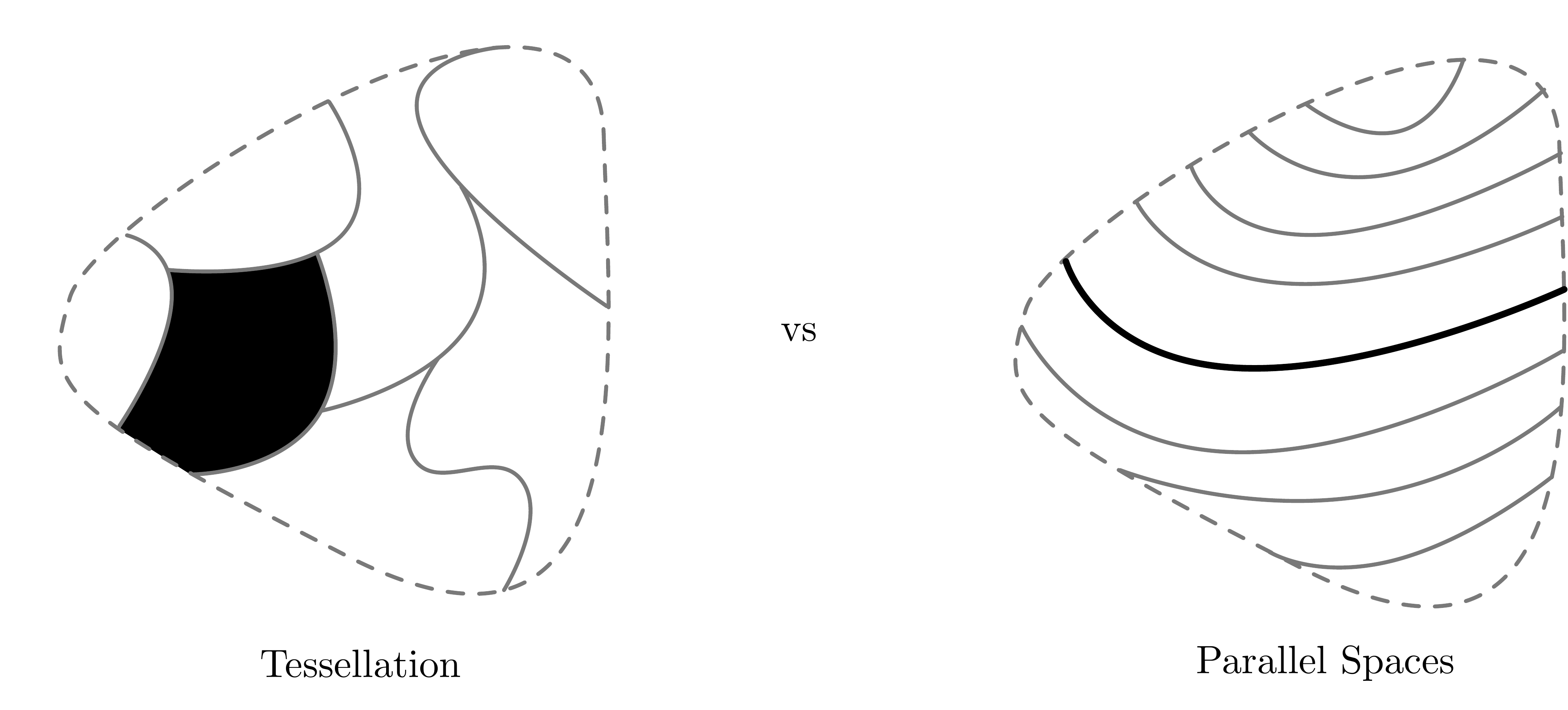}

    \caption{Comparison of tessellation and parallel spaces. The shaded regions denote the subsets we are considering in each case. A tessellation creates disjoint, smaller subsets that are of the same dimension as the original set. Parallel spaces on the other hand creates disjoint subsets that are of a lower dimension than the original set.}
    \label{fig:tessellation}
\end{figure}

A question that can be asked here is if partitioning into parallel spaces is the only way to generate subsets that allow for faster learning. Another partitioning we can consider is a tessellation, as in Figure \ref{fig:tessellation}. A tessellation partitions a space by drawing boundaries, which together with the elements that are contained within them form the subsets or partitions. Voronoi diagrams are examples of tessellations \citep{Reddy2012, Lee1982}. As with parallel spaces, a tessellation can make the search for a particular element faster, if we know which partition it belongs to. In this case, this is because the distance we would have to move from an initial guess in the partition is smaller than if we were to make an initial guess arbitrarily in the full space. We state that a tessellation considers transfer in terms of similarity. 

\begin{figure} 
    \centering
    \includegraphics[width=0.32\textwidth]{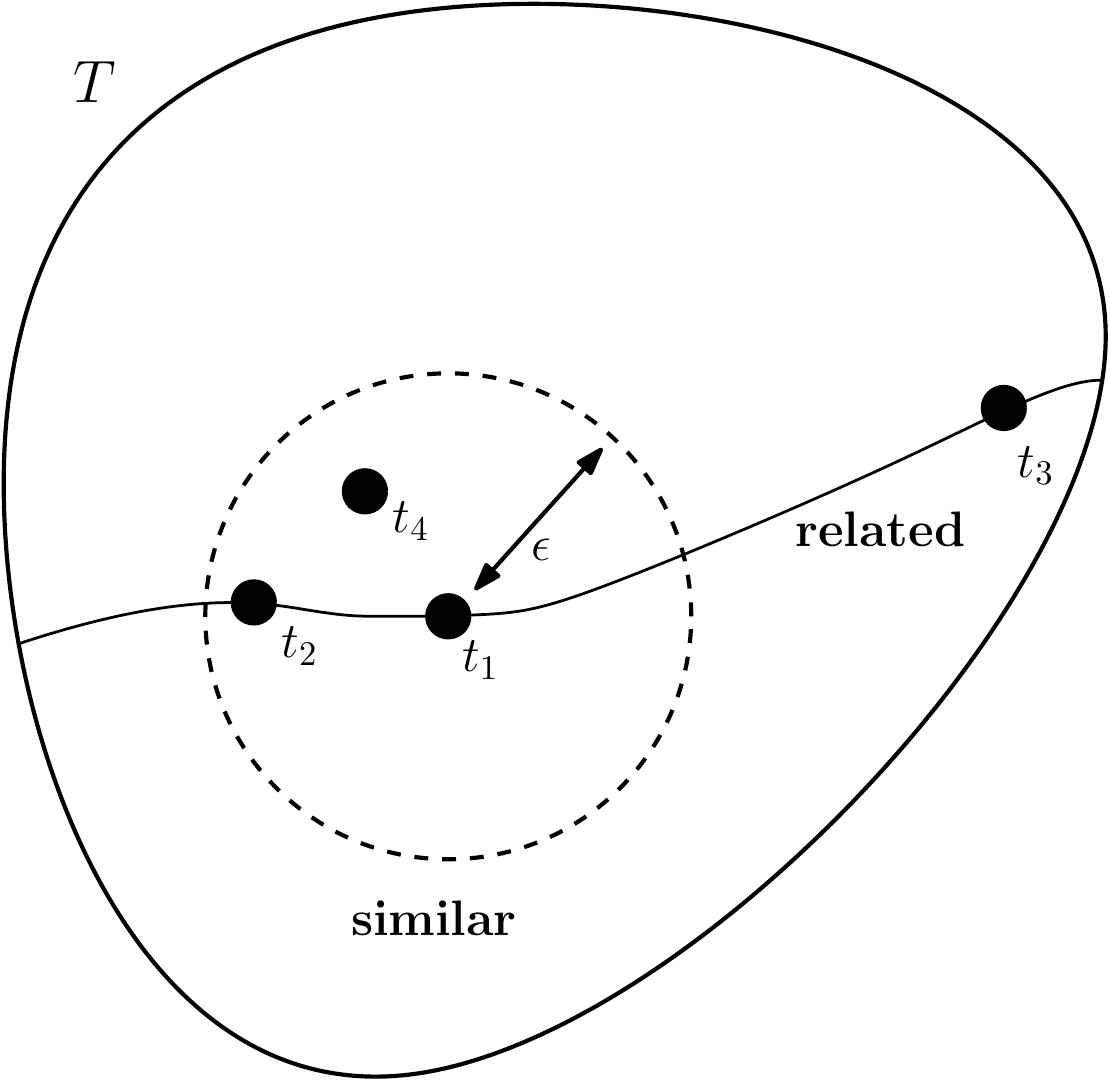}
    \caption{The distinction between similarity and relatedness. Similarity is defined in terms of an $\epsilon$-ball around a particular element. Here, points $t_2$ and $t_3$ are similar to $t_1$. On the other hand, relatedness is defined in terms of a transformation set; this relationship is shown as a line joining them. Thus, tasks $t_1, t_2$ and $t_3$ are related to each other. However, $t_4$ is not related to $t_2$ and $t_1$ though they are similar; $t_3$ is not similar to $t_2$ and $t_1$, though they are related.}
    \label{fig:related_vs_similar}
\end{figure}

What then is \emph{similarity}? Often similarity and relatedness are used interchangeably; we believe their distinction is important. We  define two tasks as being similar to each other if, under some distance metric $\rho$ \citep{Mendelson1990, Fomin1999}, the distance between them is small. That is, $f$ and $g$ are similar iff $\rho(f, g) < \epsilon$, where $\epsilon$ is to be chosen. Similarity is therefore a \emph{geometric notion}. On the other hand, two elements are related if they can be transformed from one to another, given a considered set of transformations. Relatedness is a \emph{transformative notion}. These are independent properties that can be placed on a set of elements, as illustrated in Figure \ref{fig:related_vs_similar}. 

A parallelised space is a natural image of the way that we had defined relatedness; it presents additional structure that allows for transfer. All the parallel spaces created in this way define a partitioning of the ambient space; it separates a set into a set of non-intersecting subsets that locally, have a shared, similar notion of relatedness. A tessellation then is a natural image of similarity.

Transfer with respect to similarity occurs when we update a pre-trained model using new data, with the expectation that the original pre-trained model is a good initial guess for the new task. Often this involves training the original model on a very large data set, a subset of which is expected to be useful for a new task \citep{Huh2016}. Such an approach gives us a partition of a tessellation if we consider all the elements that are a certain distance away from the pre-trained model, with respect to the loss function.  While this is a valid approach, and has shown empirical success, we focus on the use of parallel spaces in the present work for the following reason.

A partitioning gives us a global organisation scheme within a space of tasks. In relatedness, the consistency of parallel spaces depends on transformation group that is chosen. That is, if we were to pick out two tasks that are related to each other, then a third task, which is related to the second would also be related to the first, all with respect to the transformation set. On the other hand, generating a tessellation using similarity involves picking some maximum distance, with respect to the loss function, that we are willing to tolerate, as well as a countable set of reference elements with respect to which the similarity is measured. This is similar to building Voronoi diagrams for k-means clustering \citep{Reddy2012}, or k-NN methods \citep{Lee1982}. When transferring in this way, it is assumed that the pre-trained model is one such reference element\footnote{Our notion of similarity and its interpretation using tessellations can describe why catastrophic forgetting occurs \citep{Goodfellow2013}. The further away from the original pre-trained model we move, the less similar the new models are; thus, gradually, we lose any resemblance to it. In the case of transfer using relatedness, such an issue does not occur as long as we fix and stay on the parallel space we operate on.}. Thus, similarity can only be measured relative to these reference elements, and the consistency of the notion of similarity depends on the choice of such reference elements. We believe that doing this such that we can gain insight into a global organisation of the set of all considered tasks is a more difficult problem; for example, how do we pick useful reference tasks?

\begin{figure} 
    \centering
    \includegraphics[width=0.7\textwidth]{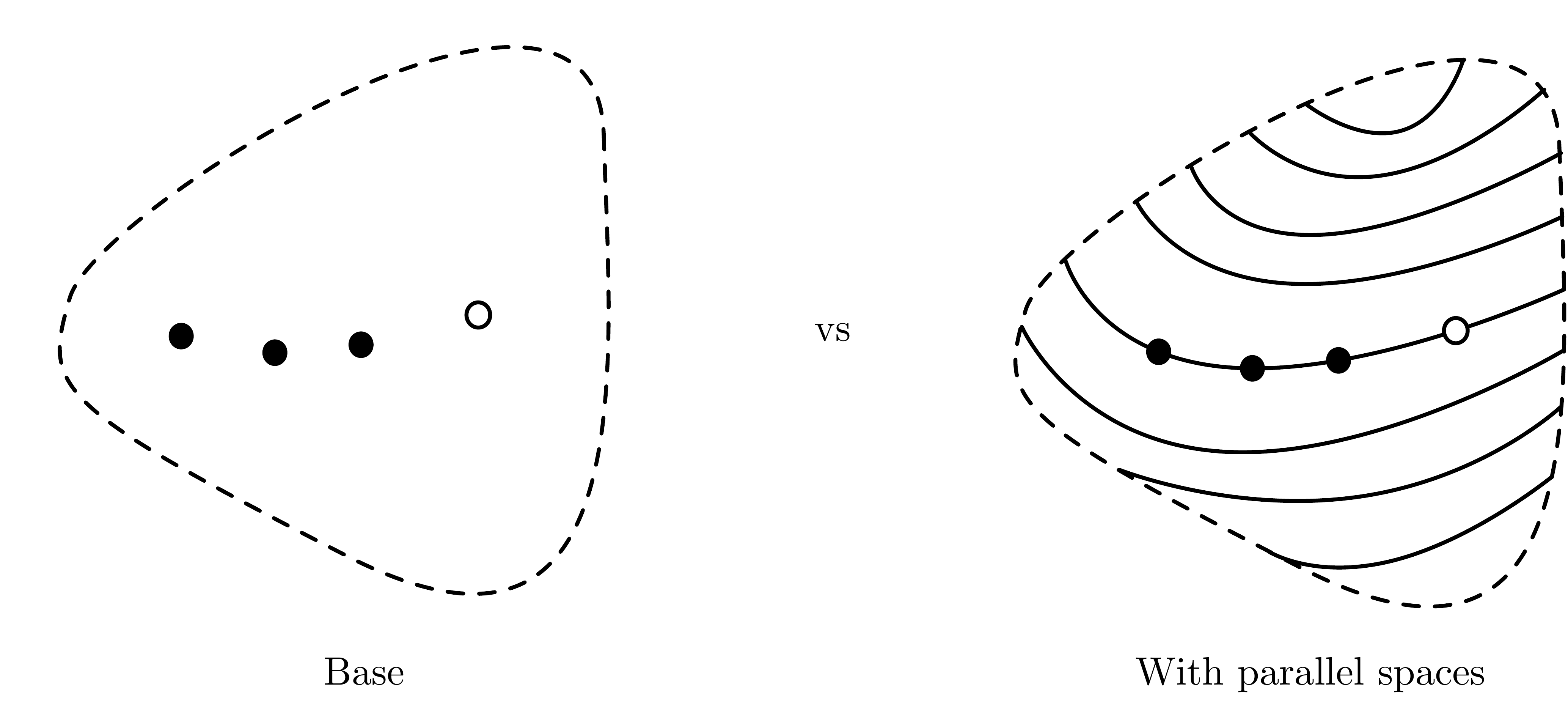}

    \caption{When carrying out transfer, we learn solutions to the original tasks, along with parallel spaces. The filled in balls represent tasks that are learned initially, while the empty ball represents a new task that is to be learned. We see in the right panel that all these tasks are related (by assumption or result), in that they lie on the same parallel space. Thus, when attempting to solve the new task, if we know that it is related to the previous tasks, we will only have to search along the parallel space they exist on.}
    \label{fig:structure}
\end{figure}

As such, when learning to carry out transfer, we conclude that it is useful to learn the solution(s) to the original task(s) such that we have access to a notion of parallel spaces of the full task space. If we train originally on a single task, then we could assume this structure; if we have multiple tasks, we can try to learn this structure using MTL. When doing the latter, we will have to make some meta-assumptions about this structure. Most state-of-the-art methods that learn in such a way, make the assumption that all tasks that are originally trained on are related, and therefore lie on the same parallel space. Then given a new task, we would assume it lies on the same parallel space, and train only along this space. This is shown in Figure \ref{fig:structure}.

It is possible to also consider a notion of a meta-objective that makes assumptions about downstream uses of the learned models. In this, the parallel spaces that tasks lie on will depend on this meta objective, and its goal is to find an organisation of space of tasks that is useful for what ever the learned models are used for. In essence, this attempts to give different scores to different ways of partitioning the original space; there is not a unique way of doing this, unless some additional criteria are specified. The meta-objective tries to specify such additional criteria.

\section{Formalism} \label{sec:formalism}
\subsection{Relatedness}
In general, two tasks of some set are related with respect to a set of transformations that can act on this set. If a task $f$ from a set of tasks $T$ is described as a function $f: X \rightarrow Y$, then a transformation is a map $\pi: T \rightarrow T$. Certain consistency and other useful properties can be achieved by assuming that this set is a group \citep{Lee2001}. We assume that the  chosen group is a Lie group, and that its action on $T$ is locally free \citep{Lawson1971, Camacho2013} and regular \citep{Olver1995}\footnote{Such assumptions do not penalise generality much, and lead to useful theoretical  guarantees; these assumptions are usually violated in what are typically considered as pathological scenarios.}. Then, given a transformation group $\Pi$ and a set of tasks $T$, tasks $f, g \in T$ are related if there exists $\pi_{fg} \in \Pi$ where $\pi_{fg}(f) = g$; since $\Pi$ is a group, this also implies that $\pi^{-1}_{fg}(g) = f$. The set of related tasks is the orbit of this group action, given by 
\begin{align}
[f]_\sim := \{\pi(f):\pi \in \Pi\}.
\end{align}
The orbit is an equivalence class that is induced by an equivalence relation $\sim$ on $T$, given by 
\begin{align}
f \sim g \iff f(x) = \pi(g)(x), \: \forall x\in X,
\end{align}
where $\pi \in \Pi$. We can then find a quotient space $T/_{\sim}$  of equivalence classes under this relation.
Since this equivalence relation is derived from the transformation group we have considered, we can  identify a particular task $f\in T$ by a representative $g$ of the set of related tasks $[f]_\sim$ it belongs to, and an appropriate transformation $\pi_{gf} \in \Pi$. 

Each task belongs to a unique equivalence class (set of related tasks); the equivalence relation partitions the space. We will see in Section \ref{sec:invariants} that the dimensionality of each equivalence class depends on the dimensionality of the transformation group. Thus, the particular partitioning we obtain from this is a parallelisation of the set of tasks, we have described previously in Section \ref{sec:relatedness}. 

We now introduce the statement of equivariance of the learning algorithm $\Alg_L$. We denote $\Model$ as the space of solutions for the learning tasks $T$ and $m_f$ as the solution we are searching for in $\Model$ for task $f$; that is, $\Alg_L(f) = m_f$. Equivariance states that if $\pi_{fg}(f) = g$, then $\pi_{fg}(m_f) = m_g$\footnote{The statement made assumes that $\pi_{fg}$ can act on the model space $\Model$ as well as the task space $T$. This is situational and impractical, since $\Model$ is often a smaller space than the space of target functions. In reality, and more generally, what we mean is that there is a homomorphism $\rho: \Pi_1 \rightarrow \Pi_{\Model}$, such that if $\Alg_L(f) = m_f$, $\Alg_L(g) = m_g$ and $\pi_{fg}(f) = g$, then $\rho \circ \pi_{fg}(m_f) = m_g$. Thus, the learning algorithm is $\rho$-equivariant.}. Therefore, the transformation of the target function induces a transformation in the model space. The solutions are related to each other in the same way that the target functions are. Therefore, the parallel spaces in the space of tasks induce a parallelisation of the space of solutions.. 

\subsection{Invariant quantities and parallel spaces} \label{sec:invariants}
We found that parallel spaces are a suitable structure that reflects a notion of relatedness for TL; it remains to answer how we can exactly write such a structure mathematically. As we will see, this is represented succinctly and precisely by the differential geometric notion of a \emph{foliation} \citep{Lee2001, Lawson1971, Camacho2013}. In this section, we will describe invariant quantities \citep{Olver1995}, as they will aide in the proceeding discussions. They will also further justify the use of parallel spaces as the additional structure of interest.  

 The definition of an invariant quantity on a set requires two components, a function, and a transformation. The function, called a \emph{quantity}, is a map $q: T \rightarrow \R^k$. Here, $T$ is the set over which the invariant quantity is to be defined, for example our set of tasks. A transformation is a map  $\pi: T \rightarrow T$. The quantity $q$ is said to be \emph{invariant} with respect to the transformation $\pi$, if
 \begin{align}
     q(f) = q\circ \pi(f),
 \end{align}
 that is,
 the value of its output for a particular $f \in T$ does not change (quantifiably) if $f$ is transformed by $\pi$.   

An example of an invariant quantity is the radius of a circle under any rotation about its centre. That is, if a point on a circle with centre $(0, 0)$ is represented in $\R^2$ by Cartesian coordinates $(x, y)$, then the quantity $r(x, y) := \sqrt{(x^2 + y^2)}$ is invariant if we transform the point $(p, q)$ using
\begin{gather}
 \begin{bmatrix} 
 	p'\\ q'
 \end{bmatrix}
 = g_x(p,q) = 
  \begin{bmatrix}
   \cos(x) & -\sin(x) \\
  	\sin(x) & \cos(x) 
   \end{bmatrix}
    \begin{bmatrix} 
 	p\\q
 \end{bmatrix},
\end{gather}
where $x$ specifies an angle. The proof of this follows intuitively. Other invariant quantities include the circumference and area. In fact, for a circle, all such quantities are functions of the radius $r$. It should be noted that these quantities are invariant with respect to a wider class of transformations, called the rigid body transformations. These are the set of rotations, reflections and translations. Of course, circles with different radii will have a different \emph{value} of the invariant quantity; however, given a particular circle, whatever its radius, this value will not change under the prescribed transformations. 

The notion of an invariant quantity is intimately connected to a transformation group, particularly to Lie groups and their actions. It is known that the number of independent, scalar valued invariant quantities of a regular group action is completely determined by the dimension of the resultant space of related elements given by the transformation group \citep{Olver1995}; that is, $k = d - n$, where $k$ is as above, $d$ is the dimension of the full space, and $n$ is the dimension of the set of related elements. In specifying a transformation group and its action, we are able to generate a parallel partitioning of the space of tasks $T$, and by the statement of equivariance, of the space of solutions $\Model$. In doing so, we have invariably specified an invariant quantity; something that stays constant within a set of related tasks.

By using invariant quantities we develop hierarchical categorisation schemes of the \emph{set of all considered tasks}. Let us revisit the pendulum problem from Section~\ref{sec:introduction}. Pendulums are classical dynamical systems; that is, from all possible dynamical systems we can conceive of, they are elements of the subset that contains dynamical systems that satisfy Hamilton's principle \citep{Arnol2013}. Then from that set, pendulums in particular are dynamical systems whose motion is restricted to a circle, and operate under a restoring force field (gravity). These particular constraints \emph{identify} what we think of conceptually as pendulums. Such constraints are reflected in the ODEs that we write for pendulums; the invariant quantities are described by the \emph{form} of the ODEs, written in the language (representation scheme) of calculus. 

If we then consider that all pendulums can be described using a model space with finite parameters, then the subset of models that describe pendulums (that differ only in mass and length) will contain some characteristic property that specifies that the elements within it are pendulums. This characteristic property is exactly given by the invariant quantity given w.r.t. the transformations that can transform one pendulum to another. The invariant quantity here, now written in a different language (the representation scheme of the model space), describes the information contained in the form of the ODEs of pendulums.  

The invariant quantity gives us a way to numerically identify a set of related tasks. Of course, it depends on the representation scheme chosen for both the space of models as well as the structure of the parallel spaces. As we will see in the next section, foliations are a natural way to describe parallelised spaces. Foliations also provide local invariant quantities in local coordinate systems; in fact, the definition of a \emph{regular} foliation is in terms of rectifying coordinates that separate the invariant quantity and the remaining degrees of freedom into complementary subspaces.

\subsection{Foliations} \label{sec:foliations}
\begin{figure} 
    \centering
    \includegraphics[width=0.65\textwidth]{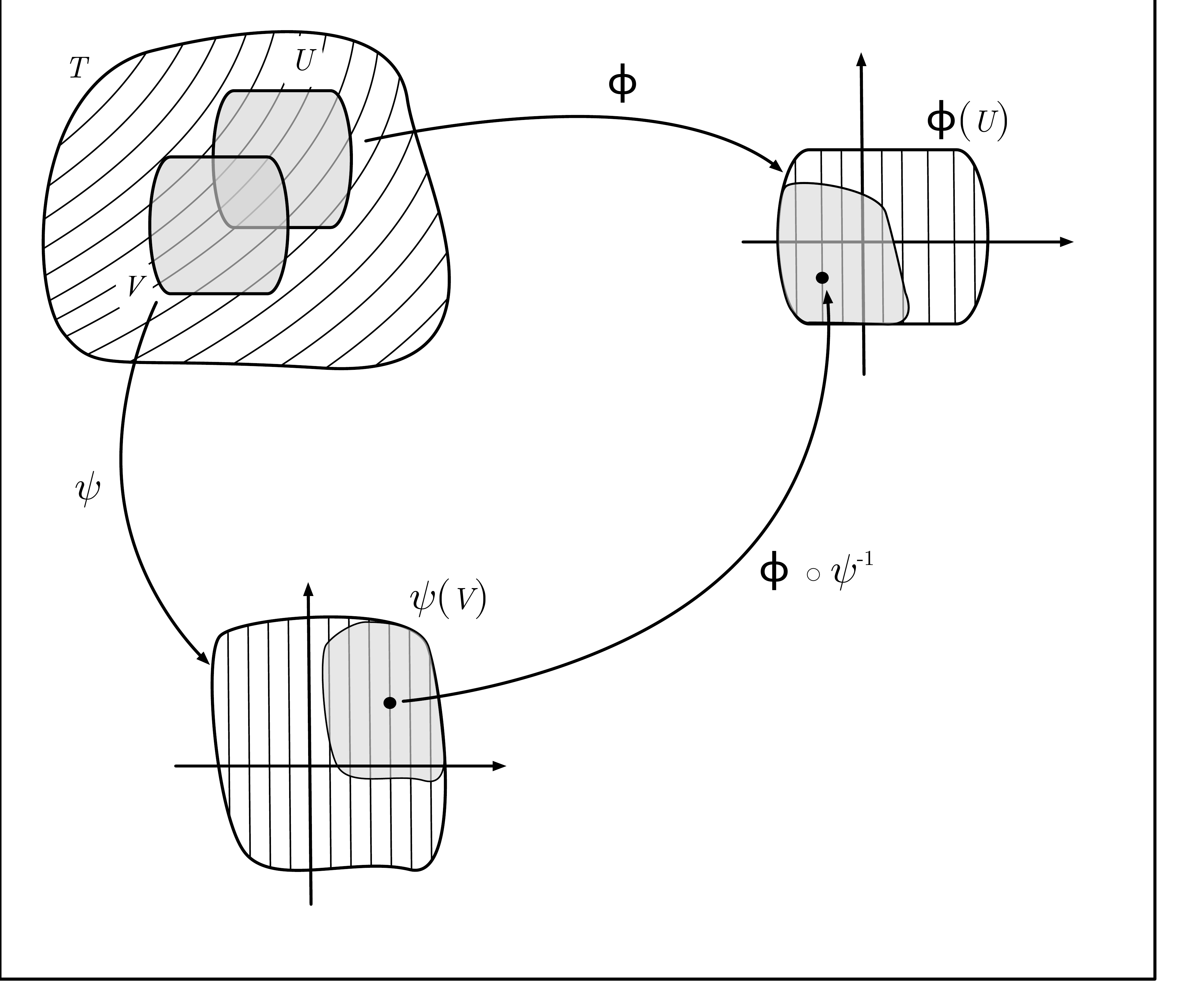}

    \caption{A graphical description of a regular foliated manifold on a manifold of dimension 2. We see 2 charts, $\phi$ and $\psi$, which map local neighbourhoods $U$ and $V$ into their respective coordinates. In the overlapping region, points on a parallel line in $\phi(U)$ are mapped onto a corresponding parallel line in $\psi(V)$. }
    \label{fig:foliation}
\end{figure}

In this section, we will briefly introduce foliations as a way of representing parallelised spaces. We refer the reader to \citep{Lee2001, Camacho2013, Lawson1971, Thurston1988, Stefan1974} for a more detailed introduction to manifolds, and regular and singular foliations.  

We will consider the spaces of tasks and models as smooth manifolds since manifolds naturally give a notion of representations and transformations between them\footnote{A smooth manifold also implies that the movement between different tasks is smooth. We consider this particular scenario for its theoretical benefits. Further, it makes intuitive sense. For example in the case of pendulums, we would expect that changing pendulums by changing masses and lengths is a smooth process, up to an extent.}. A manifold is a topological space that \emph{locally} looks Euclidean, but can be different globally. This described by realising abstract elements on a manifold using coordinate charts that map local portions of the manifold homeomorphically to a Euclidean subspace. The choice of coordinate charts is hardly unique, but the theory of differential geometry handles this constricting chart transition maps to adhere to defining consistency requirements. This allows for the key philosophy of manifolds; any theory that is developed on a manifold is \emph{independent} of the choice of charts. Charts are then choices of representations of a local part of the manifold; chart transition maps are transformations between equivalent representations.

In order to define a regular foliation, we choose charts that satisfy two properties. The first of these requires the charts map into two complementary subspaces of $\R^d$, where $d$ is the dimension of the manifold. That is,
\begin{equation}
\phi: (U \subset M) \rightarrow \big((U_\R^m \times U_\R^n) \subset \R^d \big), 	
\end{equation}
where $d = m + n$ and $M$ is the manifold. This means that that under a different chart, the same point will be mapped onto a similarly decomposed space. Secondly, we require that the transition of one of the subspaces depends only on the value of that particular subspace in the original chart. 

To see this, consider a region of the manifold $U \subset M$ on which two charts $\phi$ and $\psi$ can be applied. The chart transition is given by $h: \phi(U) \rightarrow \psi(U)$. The first condition on the chart tells us that for a point $p \in U$, where $\phi(p) = (x, y)$, $\psi(p) = h((x, y)) = ((x', y') \in U_{\R^m}' \times U_{\R^n}')$. We can decompose $h$ into $h_m$ and $h_n$, which map $U$ into  $U_{\R^m}'$ and $U_{\R^n}'$ respectively. The second condition requires that $h_m: U_\R^m \rightarrow U_{\R^m}'$, and $h_n: U_\R^m  \times U_\R^n\rightarrow U_{\R^n}'$. Thus one of the components of a point, under a transition, only depends on where in that component the point is.

The subset of the manifold that consists of a constant (up to transitions) value of this component is called a \emph{leaf}. Leaves are immersed, connected, non-intersecting submanifolds of $M$. We see immediately from Figure \ref{fig:foliation} that this precisely describes our notion of partitioning with parallel spaces. In particular, the charts, as chosen with the properties above, give us a set of rectified coordinates, where the notions of which leaf a point is on, and the where on the leaf it is are clearly separated. A foliation therefore gives us the theoretical tools to talk about parallel spaces.

There are several ways by which a foliation can be constructed \citep{Lawson1971, Camacho2013}. We have seen one when we constructed our parallel spaces using a transformation group. Generally, the orbit of a Lie group action that is locally free exactly determines a foliation that describes this partitioning. 

A question that remains is whether any arbitrary foliation can give us a notion of a set of transformations, based on which we can talk about related tasks. That is, instead of defining a foliation from a set of transformations, can we make a reverse construction? In general, it is likely that a particular leaf will have an associated Lie group that acts on it. However, this group and its action could change smoothly when moving between leaves. A further discussion of this is beyond the scope of the present work and will be left for future work.

\section{Applications in the literature and future work} \label{sec:future}
In the way that foliations are defined, we can see some immediate areas of the literature in which the notion of a foliation is used, at least implicitly. In MTL \citep{Caruana1997, Ruder2017}, we see this when carrying out hard parameter sharing, where a portion of parameters is kept consistent between tasks. Thus, we implicitly assume that these tasks lie on the same leaf, and the definition of the leaf is given by the shared parameters. In soft parameter sharing, the shared components are given free reign, but are regularised to be close to each other. In the language of foliations, this means that we allow the tasks to lie on leaves that are close to each other; that is leaves that are \emph{similar}.

In \citep{Steindor2018, Hausman2018,Janith2019}, the hard parameter sharing of MTL is used for TL in models and in policies of RL; distributions over task specific parameters are learned as latent variable. Here, once the original tasks are solved, a new task is solved by keeping the shared parameters from the original tasks fixed. That is, we carry out what we described in the right panel of Figure \ref{fig:structure}. It is possible that other work, such as neural architecture search \citep{Elsken2018}, MAML \citep{Finn2017} and other meta-learning techniques, such as \citep{Snell2017, Vinayls2016} also make use of foliations implicitly. The dissemination of this is left for future work.

In the examples above, the manifold is assumed to be $\R^d$, and we are using a trivial foliation by directly using the given rectified coordinates. Foliations are much richer than these. In future work we will be looking at incorporating foliations in general into the theory of, and solutions to TL. In addition to this, we will study how different foliations inform learning complexities of different problems.

It has also not escaped our notice that the notion of transfer, as we described it, can have possible applications in carrying out \emph{scientific discovery} using machine learning. At the expense of detail, we consider scientific discovery as the process of using empirical data to obtain an \emph{understanding} and an \emph{organisation} of different physical phenomena. The understanding can be mapped to models, and the organisation of such models can be mapped to how they relate to and inform each other. Our description of TL do just that. A possible example application we are considering is the discovery of elementary systems that can be used to build complex models for biological systems \citep{Alon2019}.

\section{Conclusion}
In this work, we described a theoretical framework by which transfer learning can be studied. In particular, we argued that transfer learning makes use of related tasks. We gave a definition by which relationships between tasks can be described in terms of transformations. Sets of related tasks partition the space of tasks into parallel spaces; these parallel spaces are a structural representation of related tasks, and are a useful tool for theoretically describing them. Foliations on differential manifolds were then found as a natural, mathematical description of parallel spaces. Finally, we briefly discussed a few areas where the notion of such foliations is used implicitly; more complicated scenarios are left for future work.

 \bibliography{main.bib}

\end{document}